\documentclass[sigconf]{acmart}
\renewcommand\footnotetextcopyrightpermission[1]{} 

\AtBeginDocument{%
  \providecommand\BibTeX{{%
    \normalfont B\kern-0.5em{\scshape i\kern-0.25em b}\kern-0.8em\TeX}}}

\setcopyright{acmcopyright}
\copyrightyear{2020}
\acmYear{2020}
\acmDOI{}

\acmConference[]{ACM WSDM}{March 8--12, 2021}{Jerusalem, Israel}
\acmBooktitle{ACM WSDM
  March 8--12, 2021, Jerusalem, Israel}
\acmPrice{}
\acmISBN{}

\begin{document}
\fancyfoot{}
\fancyhead{}

\title{Repurposing TREC-COVID Annotations to Answer the Key Questions of CORD-19}

\author{Connor T. Heaton}
\affiliation{%
  \institution{Pennsylvania State University}
  \city{State College}
  \state{PA}
  \postcode{16802}}
\email{czh5372@psu.edu}

\author{Prasenjit Mitra}
\affiliation{%
  \institution{Pennsylvania State University}
  \city{State College}
  \state{PA}
  \postcode{16802}}
\email{pum10@psu.edu}

\begin{abstract}

The novel coronavirus disease 2019 (COVID-19) began in Wuhan, China in late 2019 and to date has infected over 14M people worldwide, resulting in over 750,000 deaths\footnote{https://www.worldometers.info/coronavirus/}. On March 10, 2020 the World Health Organization (WHO) declared the outbreak a global pandemic. Many academics and researchers, not restricted to the medical domain, began publishing papers describing new discoveries. However, with the large influx of publications, it was hard for these individuals to sift through the large amount of data and make sense of the findings. The White House and a group of industry research labs, lead by the Allen Institute for AI, aggregated over 200,000 journal articles related to a variety of coronaviruses and tasked the community with answering key questions related to the corpus, releasing the dataset as CORD-19. The information retrieval (IR) community repurposed the journal articles within CORD-19 to more closely resemble a classic TREC-style competition, dubbed TREC-COVID, with human annotators providing relevancy judgements at the end of each round of competition. Seeing the related endeavors, we set out to repurpose the relevancy annotations for TREC-COVID tasks to identify journal articles in CORD-19 which are relevant to the key questions posed by CORD-19. A BioBERT model trained on this repurposed dataset prescribes relevancy annotations for CORD-19 tasks that have an overall agreement of 0.4430 with majority human annotations in terms of Cohen's kappa. We present the methodology used to construct the new dataset and describe the decision process used throughout. 


\end{abstract}


\maketitle

\section{Introduction}



The novel coronavirus disease 2019 (COVID-19) began in Wuhan, China in late 2019 and to date has infected over 14M people worldwide, resulting in over 750,000 deaths\footnote{https://www.worldometers.info/coronavirus/}. On March 10, 2020 the World Health Organization (WHO) declared the outbreak a global pandemic. Many academics and researchers, not restricted to the medical domain, began publishing papers presenting new discoveries related to COVID-19. Although well intentioned, the huge increase of publications about COVID-19 made it difficult for medical professionals to sift through the data and identify actionable insights. 

Hoping to encourage a more unified, organized investigation into the virus, the White House and a group of leading research labs in industry, lead by the Allen Institute for AI, released the CORD-19 dataset in March of 2020 \cite{Wang2020CORD19TC}. The dataset contains academic journal articles relating to a variety of coronavirus and related viral infections, not only COVID-19, sourced from PubMed Central (PMC), PubMed, the World Health Organization (WHO), bioRxiv, medRxiv, and arXiv. Furthermore, the dataset is accompanied by 10 \textit{key questions} that the community has been tasked with answering.

As a consequence of the CORD-19 dataset not indicating which journal articles are helpful in answering each of the key questions posed, many of the initial efforts
were positioned as clustering or data exploration studies \cite{sonbhadra2020target, le2020visualising, fister2020discovering}. However, some individuals have taken it upon themselves to assemble well-structured, task-specific datasets from a subset of documents in CORD-19. Two such datasets for question answering are CovidQA \cite{tang2020rapidly} and RECORD (\textbf{R}esearch \textit{E}ngine for \textbf{C}OVID \textbf{O}pen \textbf{R}esearch \textbf{D}ataset) \cite{muffo2020record}. While these datasets are a valuable resource to the community, their heavy reliance on human annotation limits their utility and scalability as each dataset contains less than 150 records. 

The information retrieval (IR) community reframed the key questions asked in CORD-19 to more closely resemble a TREC competition, calling the resulting competition TREC-COVID \cite{voorhees2020trec}. In each round of competition contestants are given a list of queries, or tasks, for which related documents are desired. Contestants then perform the queries using their proposed model, returning an ordered list of documents expected to be relevant to the query. To assess the performance of teams participating in the competition, human annotators prescribe relevancy annotations to journal articles that are returned most often for each task in TREC-COVID at the end of each round of competition. 

Although these two activities have received much attention from their respective communities, they are usually viewed in isolation. That is, efforts towards answering the queries associated with TREC-COVID have not been directly leveraged towards answering the key questions posed by CORD-19. While valuable human annotations have been obtained for TREC-COVID tasks, ground-truth labels for CORD-19 task have yet to be obtained even though it asks similar questions on the same dataset. 

Our initial attempts to train a model to perform both the TREC-COVID and CORD-19 tasks, either through multitask-learning or transfer learning, proved unfruitful. Learning to perform both sets of tasks in unison results in inferior performance than if the tasks were learned separately. 
This was an indication to us to focus more so on how annotated data can be repurposed instead of repurposing a model already trained to perform a specific task. 

We therefore present a method for re-purposing the labeled data for one task, TREC-COVID, for a task for which labeled training data is unavailable or only available in limited quantities, CORD-19. Our method begins by first defining a manual mapping from TREC-COVID tasks to CORD-19 tasks such that labels for the TREC-COVID task can be reused as labels for the corresponding CORD-19 task. We train a BioBERT model to make relevancy predictions for each CORD-19 task and compare the model's performance against that of three human annotators. We then employ a variety of techniques to refine the mapping between tasks until optimal model performance is reached. 


In total, our contributions are as follows:
\begin{itemize}
    \item Demonstrate the ability of a BioBERT model to learn to make relevancy predictions for TREC-COVID tasks, achieving a true positive rate and true negative rate of 0.75\% and 0.88\%, respectively.
    \item Present a method for repurposing the annotations from TREC-COVID towards answering the \textit{key questions} of CORD-19, achieving a Cohen's kappa of 0.443 with majority agreement human annotations.
\end{itemize}

\section{Related Literature}

\subsection{Language Modeling}

Language modeling (LM) is a long-studied discipline in Natural Language Processing (NLP) in which a model is tasked with learning the underlying distribution and relation between words in a pre-determined vocabulary \cite{jozefowicz2016exploring}. It has become common practice for a language model (LM) to first be \textit{pre-trained} on a large, general purpose corpus before being \textit{fine-tuned} for a domain-specific NLP task. Leveraging a LM in such a capacity has lead to new state-of-the-art performance in a variety of natural language understanding (NLU) and natural language inference (NLI) tasks \cite{mikolov2010recurrent, schwenk2012large, filippova2015sentence}. One of the more widely adopted pre-trained LM was released in late 2018 by Devlin et al dubbed BERT (\textbf{B}idirectional \textbf{E}ncoder \textbf{R}epresentation from \textbf{T}ransformers) \cite{devlin2018bert}. BERT was trained using two pre-training tasks: 1) Masked Language Modeling (MLM) and Next Sentence Prediction (NSP). For MLM, 15\% of tokens in the input sequence were randomly masked and BERT was asked to impute the missing tokens. In the NSP task, BERT was presented with two sentences and asked to predict whether or not the appear next to each other in a source document. 



\subsubsection{BioBERT}

Although the vanilla BERT model was shown to achieve strong performance on tasks making use of a more general vernacular, it struggled to adapt to some domain-specific applications which made use of a highly specialized vocabulary. To this end, Lee et al performed further pre-training of BERT using a corpus from the bio-medical domain, naming the new model BioBERT \cite{lee2020biobert}. Starting from the weights of the vanilla BERT model, BioBERT was further pre-trained on an 18B word corpora composed of PubMed abstracts and full-text articles from PMC. BioBERT was shown to achieve new state-of-the-art performance on a variety of NLP tasks tailored to the bio-medical domain including named entity recognition (NER) and question answering (QA) among others \cite{lee2020biobert}.


\subsection{COVID-19 and CORD-19}
Lead by the Allen Institute for AI and the White House, a consortium of industry leaders aggregated academic journal articles related to COVID-19 and other coronaviruses, releasing the dataset to the public as CORD-19 \cite{Wang2020CORD19TC}. The dataset and supplemental material are freely available on Kaggle\footnote{https://www.kaggle.com/allen-institute-for-ai/CORD-19-research-challenge}. 

Sonbhadra et al \cite{sonbhadra2020target} present a method for identifying journal articles relevant to each of the 10 CORD-19 tasks. The proposed method begins by clustering articles in CORD-19 based on their TF-IDF embeddings and training a one-class support vector machine (OCSVM) to identify samples belonging to each centroid. Next, doc2vec is used to create ebmeddings that represent each centroid and each Kaggle task. These embeddings are then used to calculate the cosine similarity, and consequently the association between, each centroid and Kaggle task. Although this method achieves high levels of performance with F1 scores approaching 0.999, the ground truth labels are assumed based on embedding vector similarity without human confirmation. It shows strong ability to identify which set of centroids a particular journal was assigned to, but whether or not the journal articles assigned to that centroid are truly relevant to the task is not confirmed. 

The National Institute of Health (NIH) also compiled a dataset of articles related to the ongoing pandemic and released to the community as \textit{LitCovid} \cite{chen2020keep}. Articles in this dataset are sourced from PubMed and exclusively discuss COVID-19. The dataset contains around 8,000 articles each annotated as one or more of the following categories: \textit{General}, \textit{Transmission}, \textit{Treatment}, \textit{Case report}, \textit{Forecasting}, \textit{Prevention}, \textit{Mechanism}, and \textit{Diagnosis}. Gutierrez et al assessed the performance of a variety of neural architectures on the dataset, including LSTM, CNN, BERT, BioBERT, and other transformer-based models \cite{gutierrez2020document}. Ultimately, the group found that BioBERT and the Longformer, a variant of the transformer architecture \cite{vaswani2017attention}, produced the best results with an micro-F1 score and accuracy of 81\% and 69\%, respectively, on the test set. BioBERT consistently outperformed the vanilla BERT models throughout their experiments, demonstrating the benefit of utilizing a LM fine-tuned for the biomedical domain.

\subsection{Transfer Learning}
As defined by Torrey et al, transfer learning (TL) ``... is the improvement of learning in a new task through the transfer of knowledge from a related task that has already been learned'' \cite{torrey2010transfer}. For a simple, pertinent example we remind the reader of the process of pre-training a LM on a large, general purpose before fine-tuning it for a specific task. In doing so, the model first learns general knowledge about words in the vocabulary being learned before refining it's understanding for optimal performance on a specific task.

TL is typically discussed with respect to the \textit{model} - \textit{i.e.} a model learning to perform task B will make use of what it learned in learning to perform task A. In our experiments, however, we utilized TL in the \textit{data space} instead of the \textit{model space}. Human annotators prescribed relevancy annotations for journal articles in CORD-19 with respect to TREC-COVID tasks, and using those labels, we re-purpose the dataset such that it can be used to train a model to perform a new task. The \textit{knowledge} obtained from the TREC-COVID annotations is never truly ``shared'' with the model learning to prescribe CORD-19 relevancy annotations, but used to re-purpose the dataset for a new task.



\subsection{Question Answering}
Although CORD-19 on it's own is a completely unstructured dataset, some individuals have taken it upon themselves to assemble well-structured, task-specific datasets from a subset of documents in CORD-19. Two such datasets are CovidQA \cite{tang2020rapidly} and RECORD (\textbf{R}esearch \textit{E}ngine for \textbf{C}OVID \textbf{O}pen \textbf{R}esearch \textbf{D}ataset) \cite{muffo2020record}. Both datasets present a model with a query and a context string, asking the model to identify the span of text in the context that most accurately responds to the query. Due to the reliance on human labor to construct these fine-grain, task-specific datasets, the datasets only contain 124 and 112 question-answer pairs, respectively. 


\section{Datasets}

\subsection{TREC-COVID}
The TREC community answered the call to action against COVID-19 by announcing the TREC-COVID competition \cite{voorhees2020trec}. Similar to many other TREC competitions, participants are given a set of queries and asked to find documents that are relevant to the query. 
In TREC-COVID, the system returns relevant articles in the CORD-19 dataset \cite{Wang2020CORD19TC}. 
In each round of the competition, participants submit a list of articles returned for each query, ordered by predicted likelihood of relevancy. 
At the end of each round, human annotators annotate the articles that were most often returned for each query,
and they are then used to score submissions. As of completion of round three of the competition, a total of 16,677 unique journal articles in CORD-19 have received a relevancy annotation with respect to one or more TREC-COVID tasks. 

TREC-COVID began with 30 tasks, or ``queries'', and adds 5 tasks for each round of the competition. Each task is expressed in three ways: 1) as a query, typically a few words in length, \textit{i.e. ``coronavirus origin''}, 2) as a question, which poses the query in a slightly longer form, \textit{i.e. ``what is the origin of COVID-19''}, and 3) as a narrative, which further expands upon the corresponding question and query, \textit{i.e. ``seeking range of information about the SARS-CoV-2 virus's origin, including its evolution, animal source, and first transmission into humans''}. The queries for the 40 tasks included in round three of the TREC-COVID challenge are presented in table \ref{Tab:TREC-COVID-TASKS}.

\begin{table}[ht!]
    \centering
    \begin{tabular}{c | l}
      \textbf{Task ID} & \textbf{Query} \\
      \hline
      1 & coronavirus origin \\
      \hline
      2 & coronavirus response to weather changes \\
      \hline
      3 & coronavirus immunity \\
      \hline
      4 & how do people die from the coronavirus \\
      \hline
      5 & animal models of COVID-19 \\
      \hline
      6 & coronavirus test rapid testing \\
      \hline
      7 & serological tests for coronavirus \\
      \hline
      8 & coronavirus under reporting \\
      \hline
      9 & coronavirus in Canada \\
      \hline
      10 & coronavirus social distancing impact \\
      \hline
      11 & coronavirus hospital rationing \\
      \hline
      12 & coronavirus quarantine \\
      \hline
      13 & how does coronavirus spread \\
      \hline
      14 & coronavirus super spreaders \\
      \hline
      15 & coronavirus outside body \\
      \hline
      16 & how long does coronavirus survive on surfaces \\
      \hline
      17 & coronavirus clinical trials \\ 
      \hline
      18 & masks prevent coronavirus \\
      \hline
      19 & what alcohol sanitizer kills coronavirus \\
      \hline
      20 & coronavirus and ACE inhibitors \\
      \hline
      21 & coronavirus mortality \\
      \hline
      22 & coronavirus heart impacts \\
      \hline
      23 & coronavirus hypertension \\
      \hline
      24 & coronavirus diabetes \\
      \hline
      25 & coronavirus biomarkers \\
      \hline
      26 & coronavirus early symptoms \\
      \hline
      27 & coronavirus asymptomatic \\
      \hline
      28 & coronavirus hydroxychloroquine \\
      \hline
      29 & coronavirus drug repurposing \\
      \hline
      30 & coronavirus remdesivir \\
      \hline
      31 & difference between coronavirus and flu \\
      \hline
      32 & coronavirus subtypes \\
      \hline
      33 & coronavirus vaccine candidates \\
      \hline
      34 & coronavirus recovery \\
      \hline
      35 & coronavirus public datasets \\
      \hline
      36 & SARS-CoV-2 spike structure \\
      \hline
      37 & SARS-CoV-2 phylogenetic analysis \\
      \hline
      38 & COVID inflammatory response \\
      \hline
      39 & COVID-19 cytokine storm \\
      \hline
      40 & coronavirus mutations \\
      \hline
    \end{tabular}
    \caption{TREC-COVID Round 3 Task Queries}
    \label{Tab:TREC-COVID-TASKS}
\end{table}

\begin{table}[ht!]
    \centering
    \begin{tabular}{c | p{6cm}}
      \textbf{Task ID} & \textbf{Key Question} \\
      \hline
      1 & What is known about transmission, incubation, and environmental stability? \\
      \hline
      2 & What do we know about COVID-19 risk factors? \\
      \hline
      3 & What do we know about virus genetics, origin, and evolution? \\
      \hline
      4 & What do we know about vaccines and therapeutics? \\
      \hline
      5 & What has been published about medical care? \\
      \hline
      6 & What do we know about non-pharmaceutical interventions? \\
      \hline
      7 & Are there geographic variations in the rate of COVID-19 spread? \\
      \hline
      8 & What do we know about diagnostics and surveillance? \\
      \hline
      9 & What has been published about ethical and social science considerations? \\
      \hline
      10 & What has been published about information sharing and inter-sectoral collaboration? \\
      \hline
    \end{tabular}
    \caption{CORD-19 Key Questions}
    \label{Tab:CORD-19-Key-Questions}
\end{table}

\subsection{CORD-19}
Organized by the White House, Allen AI, and leading research groups, this dataset contains almost 200,000 journal articles about COVID-19 and related coronaviruses, around 80,000 of which contain the full text of the article \cite{Wang2020CORD19TC}. The dataset poses 10 ``key questions'' to the community, presented in table \ref{Tab:CORD-19-Key-Questions}.

Journal articles in the CORD-19 dataset come from sources such as PubMed Central (PMC), PubMed, the World Health Organization (WHO), bioRxiv, medRxiv, and arXiv. The dataset does not exclusively focus on COVID-19 and includes journal articles relating to other viruses such as MERS, H1N1, and SARS. New articles are periodically added to the dataset as they become available. The version of the dataset used throughout this study, unless otherwise noted, was published on July 6, 2020.

\section{Methodology}

Although the CORD-19 dataset contains over 80,000 full-text academic articles and poses 10 key questions, it gives no indication as to which articles are helpful for answering the questions being asked. However, the TREC-COVID competition is asking similar questions about the same journal articles and provides human relevancy annotations for each task at the end of each round. Seeing this opportunity, the backbone of our methodology is the re-purposing of labeled data for one task, TREC-COVID, for use in a separate, related tasks for which labeled training data is unavailable or only available in limited quantity, CORD-19. 

\begin{figure}[hb]
    \centering
    \includegraphics[scale=0.25]{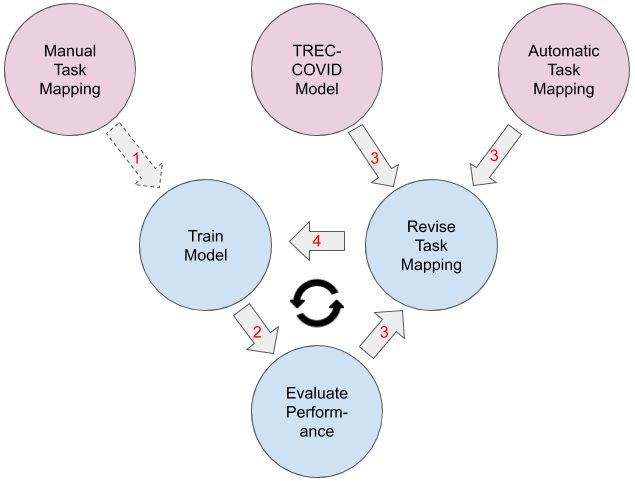}
    \caption{Workflow of our proposed method. Actions in pink nodes are only performed once at the beginning of the investigation while actions in blue nodes are iteratively performed until optimal performance is reached.}
    \label{fig:workflow}
\end{figure}

\begin{table}[!ht]
    \centering
    \begin{tabular}{c|p{5cm}}
         \textbf{CORD-19 Task ID} & \textbf{TREC-COVID Task ID}  \\
         \hline
         1 & 2, 3, 10, 13, 14, 15, 16, 17 \\
         \hline
         2 & 4, 21, 22, 23, 24, 25 \\
         \hline
         3 & 1, 2, 5, 8, 13, 15, 18, 19, 32 \\
         \hline
         4 & 1, 2, 5, 8, 13, 15, 18, 19, 28, 29, 30, 31, 32, 33, 34 \\
         \hline
         5 & 11, 17, 18, 19, 20, 28, 29, 30, 33, 34 \\
         \hline
         6 & 10, 12, 18, 34 \\
         \hline
         7 & 2, 13, 32 \\
         \hline
         8 & 6, 7, 11, 19, 25, 26 \\
         \hline
         9 & 8 \\
         \hline
         10 & 35
    \end{tabular}
    \caption{Original manual mapping from TREC-COVID tasks to CORD-19 tasks}
    \label{Tab:Manual-Task-Map}
\end{table}

We first create a manual mapping from TREC-COVID tasks to CORD-19 tasks and train a BioBERT model to predict whether or not a journal excerpt is relevant to each CORD-19 task. Next, for each CORD-19 task 20 journal articles are sampled and three human annotators gave relevancy annotations for the sampled articles and corresponding task. These annotations are used to assess model performance and the mapping between tasks is iteratively refined until optimal performance is reached. A diagram of the workflow in our methodology is presented in figure \ref{fig:workflow}.

\subsection{Manual Task Mapping}
The initial mapping between the TREC-COVID tasks and the CORD-19 tasks was done manually by one of the authors. Given the full details of both the TREC-COVID and Kaggle CORD-19 tasks, the author was asked to give their ``best guess'' judgement as to which of the TREC-COVID tasks would best align with each of the Kaggle CORD-19 tasks. The initial manual mapping was defined using task descriptions from round 3 of TREC-COVID and is presented below in table \ref{Tab:Manual-Task-Map}.

\subsubsection{Constructing Dataset}
To frame the problem as a supervised learning task, we repurpose annotations for TREC-COVID tasks to serve as ground truth labels for the corresponding CORD-19 tasks. When constructing the training set for CORD-19 task \textit{i}, we first identify the set of journal articles, $J$, that have received annotations for TREC-COVID tasks which are mapped on to CORD-19 task \textit{i}. Since the majority of CORD-19 tasks have multiple TREC-COVID tasks mapped to them and a journal article may receive an annotation for more than one TREC-COVID task, it is possible for a journal article \textit{j}, $j \in J$, to have multiple, inconsistent labels for CORD-19 task \textit{i} appropriated from corresponding TREC-COVID tasks. For example, if journal article \textit{j} was annotated as ``relevant'' for TREC-COVID task \textit{x} but ``not relevant'' for TREC-COVID task \textit{y}, and both TREC-COVID tasks \textit{x} and \textit{y} are mapped to CORD-19 task \textit{i}, \textit{j} would be labeled as both ``relevant'' and ``not relevant'' for CORD-19 task \textit{i}. If such a situation were to occur, \textit{j} would not be included in the training set for CORD-19 task \textit{i}.

Individual records in the dataset consist of either the abstract or conclusion of a journal article. If available, the title of the journal article was prepended to either the abstract or conclusion. Therefore, any individual journal articles in CORD-19 may manifest as at most two records in the repurposed dataset. 

Each item in the newly constructed dataset is accompanied by an auxiliary sentence corresponding to the \textit{key question} of the CORD-19 task for which the item represents. The input to the BioBERT model is formatted as follows: 

\begin{center}
\textit{[CLS] JOURNAL TEXT [SEP] AUX. SENTENCE [SEP]}
\end{center}
In doing so, the model is learning a task that is similar to it's NSP pre-training task.

\subsection{Human Annotation}
In order to gauge a model's ability to prescribe relevancy predictions for the Kaggle tasks, we require a set of ground truth labels to compare it's predictions against. To this end, three human annotators were asked to annotate 20 journal articles for each Kaggle task resulting in 200 total annotations each. The human annotators were not experts in the bio-medical domain and were asked to prescribe annotations with respect to the \textit{general}, \textit{medical} domain. Put differently, the text being annotated need not be \textbf{directly} related to COVID-19 for it to be annotated as \textit{relevant} if the content of the text answers the posed question. For example, a paragraph discussing the transmission of the H1N1 virus is to be marked relevant for Kaggle task 1 although it is not describing COVID-19 in particular. This was done to make the model more amenable to generalization in other viral applications and corpora.

\subsubsection{Revising Annotations}
As none of the annotators were experts in the bio-medical domain, the initial annotations were expected to contain at least some level of noise due to misunderstanding of domain terminology, Kaggle task definitions and/or scope, and requirements for a journal to be deemed \textit{relevant}. After the three annotators gave their independent annotations, they met with the project PI to discuss items that the annotators did not agree on the correct annotation. During this time the group ensured that the annotators had a common understanding of domain terminology, task descriptions, and relevancy requirements. If the annotators could not reach a consensus agreement on their own, the project PI would cast the deciding vote. Annotators were then asked to revise their original annotations with the insights from this meeting in mind. After revisions, the annotations from our three non-expert annotators had an agreement of 0.3744 in terms of Fleiss' kappa. When compared to the majority agreement annotations, the annotators had agreements of 0.6587, 0.7538, and 0.5411 in terms of Cohen's kappa.

\subsection{Refining Task Mappings}
When comparing the predictions of a BioBERT model trained using the manual mapping with the human annotations, it was clear that the model was able to identify some sort of meaningful signal in the data, but higher of levels of performance were desired. To this end we explored how the data at our disposal, including newly obtained human annotations for CORD-19 tasks, could be used to refine the manual mapping between the TREC-COVID and CORD-19 tasks. 

\subsubsection{TREC-COVID Model}
Given that we obtained 20 human annotations for each of the ten Kaggle tasks, we wanted to see if the documents that were annotated for the Kaggle tasks were also annotated for a TREC-COVID task. As of round 3 of the TREC-COVID competition, only 9 of the documents that received a human annotation for the Kaggle tasks received an annotation for any TREC-COVID task. However, to make further use of our annotations, we train another BioBERT model to prescribe COVID-TREC relevancy annotations for each of the 40 tasks in round 3 of TREC-COVID. 

Since we are not expecting this model to be highly competitive in the TREC-COVID competition, we relax the rules of the competition slightly and reframe the task as a binary instead of tertiary classification problem. Items in the TREC-COVID dataset that received an annotation of ``somewhat relevant'' were adjusted to have a label of ``relevant.'' More concretely, the BioBERT model was asked to predict if a text snippet is either ``not relevant'' or ``relevant'' for a particular TREC-COVID task. 

Our TREC-COVID BioBERT model is then applied to each of the 200 journal articles that received human annotations for the Kaggle tasks, making relevancy predictions for each of the 40 TREC-COVID tasks. A relevancy prediction less 0.5 was taken as ``not relevant'' while any prediction greater than or equal to 0.5 was taken as ``relevant.'' These relevancy predictions for the TREC-COVID tasks can then be viewed with respect to the human relevancy predictions for the Kaggle tasks, revising the mapping between the two sets of tasks accordingly. 

\subsubsection{Automatic Task Mapping}

To further refine the manually prescribed task mappings, we explore a variety of options for automatically mapping tasks in TREC-COVID to those in CORD-19 based on different data sources and data representations. In one method of obtaining automatic mappings, BioBERT was used to create an embedding vector for each task in both TREC-COVID and CORD-19. The similarity of each TREC-COVID task to each CORD-19 task was measured using cosine similarity. 

Additionally, term frequency vectors were created to represent each of the TREC-COVID and CORD-19 tasks. Articles marked as relevant to each task were included in the construction of each embedding vector, with ngrams up to size n=5 considered. Again, the similarity of each TREC-COVID task to each CORD-19 task was measured using cosine similarity. 

\section{Experiments}
In all experiments a \textit{BioBERT-base-cased} model was fine-tuned using an Adam optimizer with lr=$5e^{-6}$, $\beta_1=0.9$, and $\beta_2=0.999$. Models trained to perform the CORD-19 tasks were trained for 20 epochs while models performing the TREC-COVID tasks were trained for only 10 epochs. In both cases, a decaying learning rate with a 10\% warmup was employed.

\subsection{Manual Mapping}
We begin experiments by fine-tuning a BioBERT model to make relevancy predictions for each CORD-19 task using original manual mapping between tasks, as presented in table \ref{Tab:Manual-Task-Map}. The model had optimal agreement with the majority annotations with a relevancy threshold of 50\%. The model's performance, described in terms of Cohen's kappa, is presented below in figure \ref{fig:manual_vs_majority}.

\begin{figure}[]
    \centering
    \includegraphics[scale=0.5]{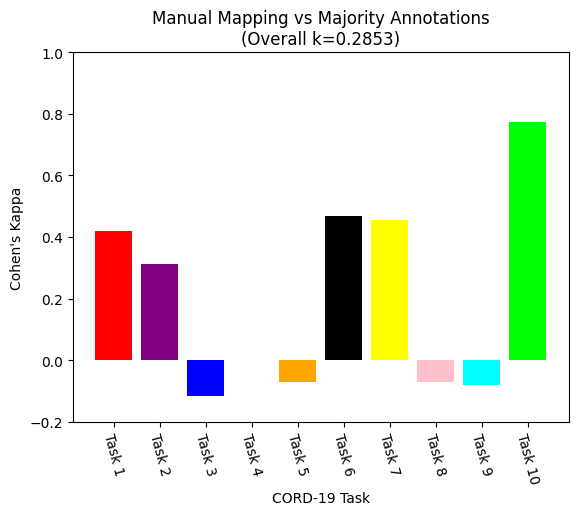}
    \caption{Performance of a BioBERT model trained to make relevancy predictions for CORD-19 tasks using the originally defined manual mapping between tasks.}
    \label{fig:manual_vs_majority}
\end{figure}

\subsection{TREC-COVID Model}

\begin{table}[]
    \centering
    \begin{tabular}{c|c|c||c|c|c}
        \textbf{Task} & \textbf{TNR} & \textbf{TPR} & \textbf{Task} & \textbf{TNR} & \textbf{TPR} \\
        \hline
        \textbf{1} & 0.78 & 0.65 & \textbf{21} & 0.90 & 0.86 \\
        \hline
        \textbf{2} & 0.95 & 0.92 & \textbf{22} & 0.91 & 1.00 \\
        \hline
        \textbf{3} & 0.82 & 0.71 & \textbf{23} & 0.89 & 1.00 \\
        \hline
        \textbf{4} & 0.92 & 0.58 & \textbf{24} & 0.93 & 0.60 \\
        \hline
        \textbf{5} & 0.87 & 0.58 & \textbf{25} & 0.96 & 0.94 \\
        \hline
        \textbf{6} & 0.77 & 0.92 & \textbf{26} & 0.88 & 0.82 \\
        \hline
        \textbf{7} & 0.93 & 0.75 & \textbf{27} & 0.79 & 0.80 \\
        \hline
        \textbf{8} & 0.84 & 0.68 & \textbf{28} & 0.91 & 0.64 \\
        \hline
        \textbf{9} & 0.93 & 0.00 & \textbf{29} & 0.78 & 0.96 \\
        \hline
        \textbf{10} & 0.77 & 0.78 & \textbf{30} & 0.87 & 0.71 \\
        \hline
        \textbf{11} & 0.86 & 0.50 & \textbf{31} & 0.86 & 0.08 \\
        \hline
        \textbf{12} & 0.56 & 0.68 & \textbf{32} & 0.94 & 1.00 \\
        \hline
        \textbf{13} & 0.80 & 0.82 & \textbf{33} & 0.92 & 1.00 \\
        \hline
        \textbf{14} & 0.90 & 0.67 & \textbf{34} & 0.97 & 0.00 \\
        \hline
        \textbf{15} & 0.88 & 0.80 & \textbf{35} & 0.96 & 0.62 \\
        \hline
        \textbf{16} & 0.89 & 0.62 & \textbf{36} & 0.83 & 0.70 \\
        \hline
        \textbf{17} & 0.85 & 0.62 & \textbf{37} & 1.00 & 0.86 \\
        \hline
        \textbf{18} & 0.77 & 0.79 & \textbf{38} & 0.80 & 1.00 \\
        \hline
        \textbf{19} & 0.94 & 0.57 & \textbf{39} & 0.67 & 0.86 \\
        \hline
        \textbf{20} & 0.78 & 0.74 & \textbf{40} & 0.40 & 0.89 \\
        \hline
    \end{tabular}
    \caption{TREC-COVID Classification Performance. Across all tasks the model achieves a TNR of 0.88 and TPR of 0.75.}
    \label{Tab:TREC-COVID-Perf}
\end{table}

When fine-tuning the BioBERT model to make relevancy predictions for the TREC-COVID dataset, the incomplete nature of the annotations necessitate the problem being framed as a binary, \textit{one-vs-all} classification problem. The model was trained using annotations for round 3 of the TREC-COVID competition and the resulting performance is described in table \ref{Tab:TREC-COVID-Perf}.

Once trained, the TREC-COVID model is then applied to the journal articles in CORD-19 that were manually annotated for the CORD-19 tasks. For each article, the model makes 40 binary predictions as to the relevancy of the article to each of the TREC-COVID tasks. For an article to be considered relevant for a particular TREC-COVID task, the model must be no less than 50\% confident the article excerpt is relevant. 

Next, for each TREC-COVID task we identify the set of journal excerpts, \textit{T}, that the TREC-COVID model believes are relevant to that task. Then, for each CORD-19 task \textit{i}, we calculate the number of relevant annotations minus the number of not-relevant annotations for that task \textit{i}. The results of this process are presented below in table \ref{Tab:TREC-COVID-Apply}. 

\begin{table}[]
    \centering
    \begin{tabular}{c|c|c|c|c|c|c|c|c|c|c}
        \textbf{CORD} $\rightarrow$ \\ \textbf{TREC} $\downarrow$  & \textbf{1} & \textbf{2} & \textbf{3} & \textbf{4} & \textbf{5} & \textbf{6} & \textbf{7} & \textbf{8} & \textbf{9} & \textbf{10} \\ 
        \hline
        1 & -1 & X & 3 & X & -3 & -3 & X & -6 & -2 & -3 \\ 
        \hline
        2 & X & X & X & X & -6 & X & X & -3 & X & 1 \\ 
        \hline
        3 & X & -3 & X & X & X & X & -6 & -6 & X & -3 \\ 
        \hline
        4 & -6 & X & X & -3 & X & X & 3 & X & -3 & -3 \\ 
        \hline
        5 & X & X & 3 & X & X & X & -3 & X & -3 & -6 \\ 
        \hline
        6 & -3 & -6 & X & -5 & -1 & -3 & 0 & -7 & -9 & -16 \\ 
        \hline
        7 & X & X & X & -3 & X & X & X & X & X & X \\ 
        \hline
        8 & -5 & -1 & -10 & X & X & -2 & 3 & X & X & X \\ 
        \hline
        9 & -5 & X & X & X & -3 & 1 & X & X & X & X \\ 
        \hline
        10 & -3 & X & -6 & -6 & -6 & 4 & X & X & X & -1 \\ 
        \hline
        11 & -3 & X & X & X & X & -3 & X & X & X & -3 \\ 
        \hline
        12 & -3 & -3 & -6 & -10 & -4 & 0 & -3 & -6 & X & -1 \\ 
        \hline
        13 & X & X & -6 & -3 & -7 & X & -4 & 1 & -6 & X \\ 
        \hline
        14 & X & X & -3 & X & X & 0 & X & -3 & X & -3 \\ 
        \hline
        15 & X & X & X & -3 & -1 & X & -3 & 1 & -3 & X \\ 
        \hline
        16 & X & X & X & -3 & X & X & X & -3 & -3 & X \\ 
        \hline
        17 & -8 & -1 & X & -3 & -1 & -3 & -3 & -1 & -6 & X \\ 
        \hline
        18 & -2 & -6 & X & X & -7 & -1 & -3 & -10 & -3 & X \\ 
        \hline
        19 & X & X & -3 & X & X & X & -2 & X & X & X \\ 
        \hline
        20 & -9 & -6 & -3 & -6 & -4 & 0 & -3 & X & -12 & -6 \\ 
        \hline
        21 & -6 & -2 & X & X & X & -6 & X & -1 & -3 & X \\ 
        \hline
        22 & -3 & X & X & -6 & X & X & X & X & X & X \\ 
        \hline
        23 & -3 & X & X & X & -3 & -3 & X & X & -3 & -3 \\ 
        \hline
        24 & -3 & X & X & X & -3 & X & X & X & X & X \\ 
        \hline
        25 & X & X & X & X & X & -3 & X & -3 & -3 & X \\ 
        \hline
        26 & X & -4 & X & X & -1 & X & -1 & X & -9 & -3 \\ 
        \hline
        27 & -4 & X & -5 & -6 & X & -3 & -4 & -7 & -12 & -7 \\ 
        \hline
        28 & X & -3 & X & X & -1 & X & X & -3 & -3 & X \\ 
        \hline
        29 & -3 & X & 3 & X & X & X & -12 & -3 & -6 & -3 \\ 
        \hline
        30 & X & X & 3 & X & X & X & -3 & -3 & -3 & X \\ 
        \hline
        31 & X & -4 & -6 & X & -3 & X & X & X & -6 & -1 \\ 
        \hline
        32 & X & X & X & X & X & -3 & X & X & X & -3 \\ 
        \hline
        33 & X & X & X & X & X & X & X & X & X & X \\ 
        \hline
        34 & X & X & X & X & X & X & X & X & X & X \\ 
        \hline
        35 & X & -1 & -1 & X & X & X & X & X & X & 1 \\ 
        \hline
        36 & X & -6 & 4 & X & X & X & -9 & X & X & X \\ 
        \hline
        37 & X & X & X & -3 & -3 & -3 & X & -3 & X & -3 \\ 
        \hline
        38 & -6 & -6 & -3 & -9 & 0 & -5 & -6 & -6 & -21 & -6 \\ 
        \hline
        39 & -3 & -3 & X & -3 & X & -3 & -6 & -6 & -16 & -3 \\ 
        \hline
        40 & -2 & X & 4 & 0 & -3 & -3 & -6 & -6 & X & -3 \\ 
        \hline
    \end{tabular}
    \caption{\textit{\# CORD-19 Relevant Annotations} - \textit{\# CORD-19 Not-relevant Annotations} for journal excerpts the TREC-COVID model believes are relevant to each of the TREC-COVID tasks. If a cell contains ``X'', the model does not believe that any journal articles annotated for the corresponding CORD-19 task are relevant to the corresponding TREC-COVID task.}
    \label{Tab:TREC-COVID-Apply}
\end{table}

\subsection{Automatic Task Mapping}
As mentioned above, we explored the use of BioBERT and term frequency embeddings when automatically mapping TREC-COVID tasks to CORD-19 tasks. The BioBERT model provided by Lee et al is used to create the BioBERT embeddings \cite{lee2020biobert}. When using BioBERT embeddings, one input sequence was generated for each task in both TREC-COVID and CORD-19. When constructing the input for TREC-COVID tasks, the input is the task \textit{query} and for CORD-19 tasks the input is the \textit{key question} for each task. The results of our automatic task mapping method using BioBERT embeddings are presented below in figure \ref{fig:biobert-mapping}.

\begin{figure*}[!ht]
    \centering
    \includegraphics[scale=0.2]{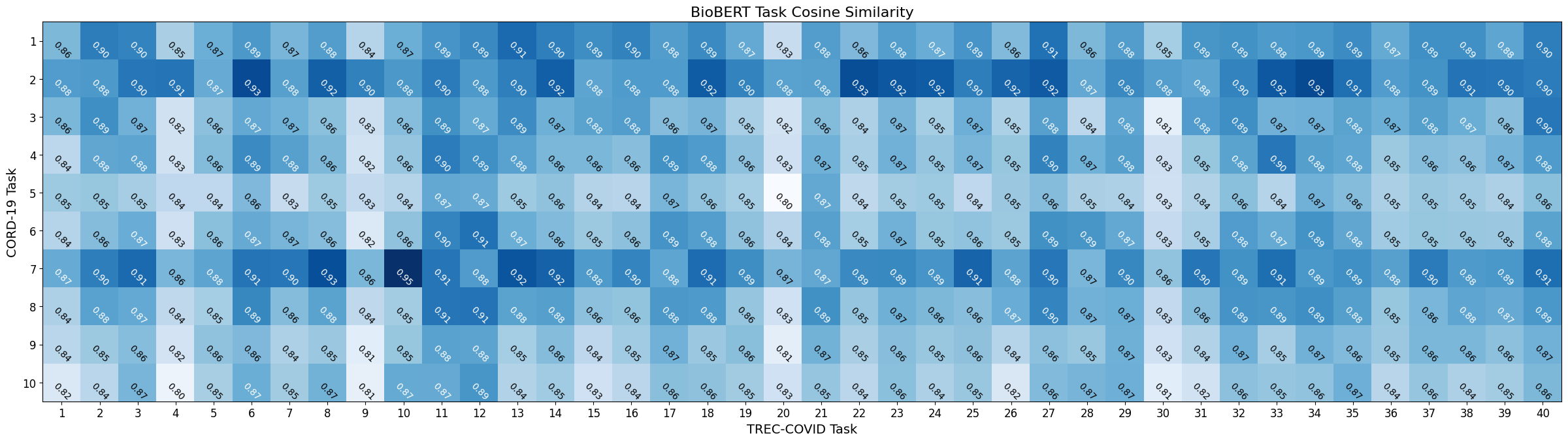}
    \caption{Automatic task mapping using BioBERT embeddings presented in terms of cosine similarity.}
    \label{fig:biobert-mapping}
\end{figure*}

\begin{figure*}[!ht]
    \centering
    \includegraphics[scale=0.2]{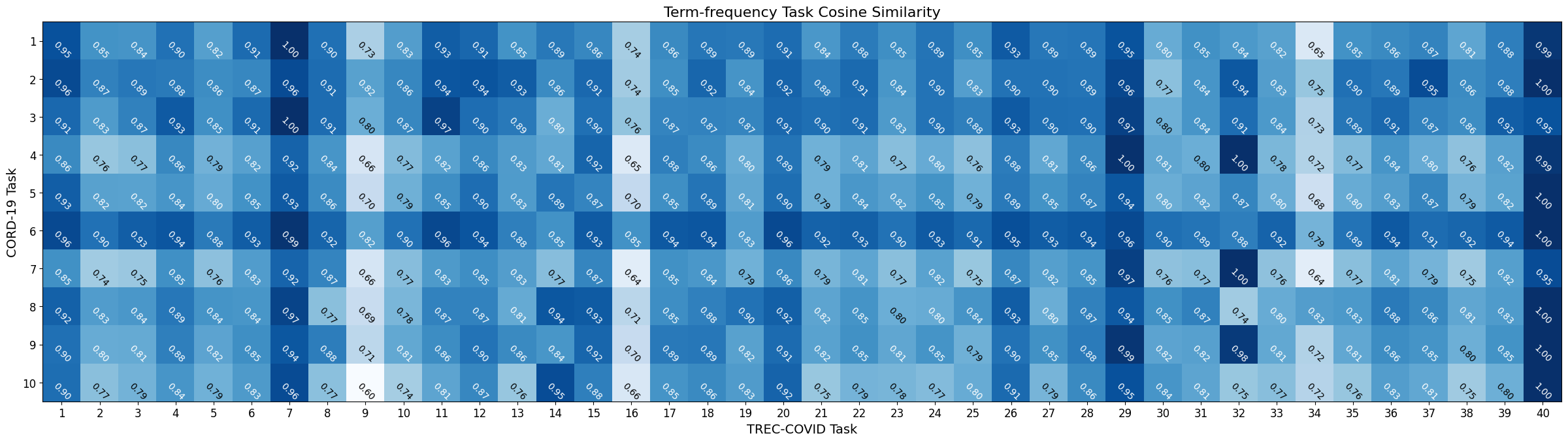}
    \caption{Automatic task mapping using term-frequency-based embeddings presented in terms of cosine similarity.}
    \label{fig:term-freq-mapping}
\end{figure*}

When using term-frequency-based embeddings to calculate an automatic mapping between tasks, we are not restricted by BioBERT's maximum input sequence length, so we can construct an embedding using more information for each task. We therefore create term-frequency embeddings that make use of articles annotated as relevant for each task when constructing the embeddings. As mentioned above, ngrams up to size n=5 are included. The results of the term-frequency-based automatic task mappings are presented below in figure \ref{fig:term-freq-mapping}. In the figure, similarity scores are normalized with respect to each CORD-19 task.

\subsection{Optimal Task Mapping}
\begin{figure}[]
    \centering
    \includegraphics[scale=0.5]{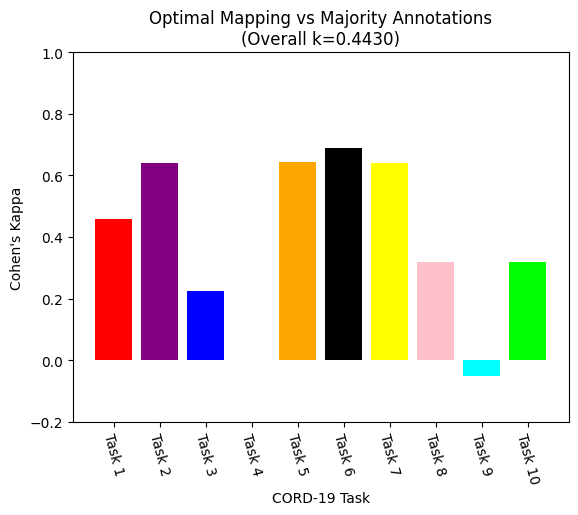}
    \caption{Performance of a BioBERT model trained to make relevancy predictions for CORD-19 tasks using the optimal mappings between the two sets of tasks.}
    \label{fig:optimal_vs_majority}
\end{figure}

Using the insights gleamed from applying the TREC-COVID model and the different methods to automatically map TREC-COVID task to CORD-19 tasks, we adjusted the originally defined task mapping. The mapping that lead to optimal performance is described in table \ref{Tab:Optimal-Task-Map} and corresponding performance metrics are presented in figure \ref{fig:optimal_vs_majority}. The decision process used when adjusting the mapping will be further explained in the \textit{Discussion} sections.

\begin{table}[]
    \centering
    \begin{tabular}{p{1.5cm}|p{5cm}}
         \multicolumn{1}{c|}{\textbf{CORD-19 Task ID}} & \textbf{TREC-COVID Task ID}  \\
         \hline
         \multicolumn{1}{c|}{1} & 2, 3, 10, 13, 14, 15, 19 \\
         \hline
         \multicolumn{1}{c|}{2} & 4, 22, 23, 24, 25 \\
         \hline
         \multicolumn{1}{c|}{3} & 1, 5, 17, 18, 29, 30, 36, 40 \\
         \hline
         \multicolumn{1}{c|}{4} & 1, 2, 5, 18, 28, 29, 30, 32, 33, 34 \\
         \hline
         \multicolumn{1}{c|}{5} & 11, 17, 22, 30, 33, 34, 38 \\
         \hline
         \multicolumn{1}{c|}{6} & 10, 12, 14, 18 \\
         \hline
         \multicolumn{1}{c|}{7} & 2, 4, 6, 8, 9 \\
         \hline
         \multicolumn{1}{c|}{8} & 7, 13, 15, 19, 25 \\
         \hline
         \multicolumn{1}{c|}{9} & 8, 12 \\
         \hline
         \multicolumn{1}{c|}{10} & 2, 10, 35 \\
         \hline
    \end{tabular}
    \caption{Optimal mapping from TREC-COVID tasks to CORD-19 tasks}
    \label{Tab:Optimal-Task-Map}
\end{table}

\subsection{Performance Assessment}
Both the CORD-19 and TREC-COVID competitions were active during our experiments, releasing new versions of their respective datasets as more journal articles become available. The majority of our experiments were performed during round 3 of the TREC-COVID competition, but we wanted to assess the impact of additional annotations with each round of competition. We therefore trained a BioBERT model to prescribe CORD-19 relevancy annotations using the identified optimal task mapping based on data from rounds 1-4 of the TREC-COVID competition. Furthermore, we trained a vanilla BERT model to perform this task as well to assess the performance that further in-domain pre-training had on resulting model performance. The performance achieved from performing the experiments described above are presented in table \ref{Tab:Impact} in terms of Cohen's kappa.

\begin{table}[]
    \centering
    \begin{tabular}{c|c|c|c}
        \textbf{TREC-COVID Rnd} & \textbf{N Annotations} & \textbf{BioBERT} & \textbf{BERT} \\
         \hline
         1 & 8,691 & 0.413 & 0.2636 \\
         \hline
         2 & 12,037 & 0.4412 & 0.2819 \\
         \hline
         3 & 33,068 & 0.4277 & 0.3187 \\
         \hline
         4 & 46,203 & 0.3853 & 0.3452 \\
        \hline
    \end{tabular}
    \caption{Performance of BioBERT and BERT models in terms of Cohen's kappa across rounds of TREC-COVID.}
    \label{Tab:Impact}
\end{table}

\section{Discussion}
In consulting figure \ref{fig:manual_vs_majority} we see that a BioBERT model trained to make relevancy annotations for CORD-19 tasks using the manually defined task mapping shows promising performance. The model shows a strong ability to make accurate annotations for CORD-19 task 10, but the same cannot be said for tasks 3, 4, 5, 8, and 9 - half of the ten CORD-19 tasks. When compared against majority annotations, the model has an agreement of 0.2853 in terms of Cohen's kappa. 

To squeeze more utility out of the human annotations for CORD-19 tasks, we consulted the results from applying the TREC-COVID model to journal articles which received human annotation. In doing so we are able to gain a better understanding of not only which TREC-COVID tasks correlate \textit{positively} with CORD-19 tasks, but also those that correlate \textit{negatively} with CORD-19 tasks. For example, TREC-COVID task 30 (\textit{``Is remdesivir an effective treatment for COVID-19?''}) was not mapped to CORD-19 task 3 in the manual mapping. However, in table \ref{Tab:TREC-COVID-Apply} we see that journal articles our TREC-COVID model thinks are relevant to TREC-COVID task 30 receive three \textit{more} ``relevant'' than ``not relevant'' human annotations for CORD-19 task 3, suggesting TREC-COVID task 30 should be mapped to CORD-19 task 3. Furthermore, journal articles that our TREC-COVID model thinks are relevant to TREC-COVID task 17 receive eight \textit{fewer} ``relevant'' than ``not relevant'' human annotations for CORD-19 task 1, suggesting the mapping should be removed. We use these findings to adjust the mapping between tasks accordingly if the suggested adjustment passes a simple human \textit{sanity-check}.

Turning the results of the automatic mapping methods to actionable insights required a larger degree of human intervention. Both of the methods appear capable of identifying seemingly sound correlations between two sets of tasks, but upon further inspection, some suggested mappings do not make logical sense. For example, TREC-COVID task 12 had the highest cosine similarity, 0.88, with CORD-19 task 9 based on BioBERT embeddings. This TREC-COVID task was not manually mapped to CORD-19 task 19 in the original manual mappings, but was in the optimal mappings. However, TREC-COVID task 6 had the highest cosine similarity, 0.93 for CORD-19 task 2. When looking at table \ref{Tab:TREC-COVID-Apply} we see that journal articles our TREC-COVID model believes are relevant to TREC-COVID task 6 receive six \textit{fewer} ``relevant'' than ``not relevant'' human annotations for CORD-19 task 2.

When trained using the optimal mapping from TREC-COVID tasks to CORD-19 tasks, our BioBERT model had an agreement of 0.4430 with majority annotations in terms of Cohen's kappa. In comparison with the original mapping, annotator agreement improved for all CORD-19 tasks except for tasks 4 and 10. 

In looking at table \ref{Tab:Impact} we see that the performance of a BioBERT model is relatively constant regardless of which round of TREC-COVID data was used to construct the training set. This table also shows the performance improvement which can be realized by pre-training a LM in the domain in which it will be deployed, with the vanilla BERT model consistently performing worse than BioBERT.


\section{Conclusion}
Facing a global pandemic, researchers and medical professionals around the world have organized to fight against COVID-19. Although well intentioned, a huge surge in publications made it difficult for health experts to sift through the data and turn new discoveries into actionable insights. Multiple communities assumed the burden of alleviating this issue, constructing datasets and organizing competitions, but each community had a slightly different approach to doing so. Although all communities were working towards a shared goal, artifacts resulting from one community are not always fully utilized by others, with one example being the human relevancy annotations from TREC-COVID. In our method presented above we demonstrate how the annotations for TREC-COVID tasks can be repurposed to answer the key questions being asked by CORD-19. 

\bibliographystyle{acm}
\bibliography{sample-base}

\end{document}